\documentclass{article}

    \PassOptionsToPackage{numbers, compress}{natbib}
 \usepackage[preprint]{arxiv}

\usepackage[utf8]{inputenc} 
\usepackage[T1]{fontenc}    
\usepackage{hyperref}       
\usepackage{url}            
\usepackage{booktabs}       
\usepackage{amsfonts}       
\usepackage{nicefrac}       
\usepackage{microtype}      
\usepackage{xcolor}         
\usepackage{colortbl}

\usepackage{algorithm}
\usepackage{algpseudocode}
\usepackage{graphicx}
\usepackage{amsmath}
\usepackage{multirow}
\usepackage{subcaption}
\usepackage{tcolorbox}
\usepackage{makecell}
\usepackage{wrapfig}
\usepackage{booktabs}
\usepackage{caption}
\usepackage{enumitem}

\definecolor{negcol}{RGB}{255,235,235}   
\definecolor{basecol}{RGB}{245,245,245}  
\definecolor{poscol}{RGB}{235,245,255}   

\title{Cross-Family Universality of Behavioral Axes via Anchor-Projected Representations}

\author{%
  Su-Hyeon Kim \\
  Department of Artificial Intelligence \\
  Yonsei University \\
  Seoul, South Korea \\
  \texttt{suhyeon.kim@yonsei.ac.kr} \\
  \And
  Yo-Sub Han \\
  Department of Computer Science \\
  Yonsei University \\
  Seoul, South Korea \\
  \texttt{emmous@yonsei.ac.kr} \\
}

\begin{document}

\maketitle

\begin{abstract}
Large language models from different families use different hidden dimensions, tokenizers, and training procedures, making behavioral directions difficult to compare or transfer across models. 
We introduce an anchor-projection framework that maps hidden representations from each model into a shared anchor coordinate space (ACS). 
Behavioral directions extracted from source models are projected into ACS and averaged into a canonical direction. 
For a new model, the canonical direction is reconstructed into its native hidden space using only anchor activations, without fine-tuning or target-specific direction extraction. 
We evaluate five instruction-tuned model families and ten behavioral axes. 
We find that same-axis directions align tightly across the Llama--Qwen--Mistral--Phi (LQMP) cluster in ACS. 
This shared structure transfers to downstream tasks. For the aligned LQMP cluster, held-out targets achieve \(0.83\) ten-way detection accuracy and \(0.95\) mean binary AUROC, while canonical steering induces refusal-rate shifts of up to \(\Delta = +0.46\) under distribution shift. 
Sensitivity analyses show that two source models and small anchor pools already suffice to approximate transferable directions. 
Overall, ACS provides a novel perspective on cross-family interpretability, revealing that representation-level transfer remains robust across model families. 

\end{abstract}

\section{Introduction}

Understanding and controlling the internal mechanisms of large language models (LLMs) has become central to interpretability, safety, and reliable deployment. 
High-level behaviors such as refusal, sentiment, bias, and reasoning style appear to be linearly decodable or steerable in hidden space. 
The LLM ecosystem is also rapidly diversifying, with new model families differing in architecture, hidden dimension, tokenizer, and distribution. 
This raises a practical question. When a new model is released without behavior-labeled data, can we reuse behavioral directions extracted from existing models, or must each model be analyzed from scratch?

Prior work suggests that LLM representations contain reusable structure, but mostly through within-family studies, feature-level analyses, or broad convergence hypotheses \citep{bricken2023monosemanticity, templeton2024scaling, park2024linear, huh2024platonic}. 
Other works extract behavioral steering vectors from axis-defining examples, but these vectors are defined in a target model's native hidden space and do not directly transfer across model families \citep{subramani2022steering, marks2024geometry, panickssery2024caa, arditi2024refusal}. 
What remains missing is a concrete coordinate system in which behavioral-axis directions from different model families can be compared and transferred to a previously unseen model.

We introduce an anchor-projection framework for this setting, where a fixed pool of anchor prompts serves as shared semantic landmarks across models. 
For each model, we collect hidden activations on these anchors and define an \emph{Anchor Coordinate Space} (ACS) by mapping any hidden vector to its cosine-similarity fingerprint with respect to the anchors. 
Thus, although models may differ in hidden dimension and basis, their representations can be expressed in a common coordinate system.

Given this shared space, behavioral directions extracted from source models can be projected into ACS and combined into a single canonical direction. 
If a behavior is represented consistently across families, these projected directions align and their average captures a transferable representation. 
For a new unseen model, we construct its anchor-based coordinates and reconstruct the canonical direction back into its native hidden space using only anchor activations. 
No axis-labeled data, target-specific direction extraction, or fine-tuning is required.

We evaluate this framework on five instruction-tuned model families---Llama, Qwen, Mistral, Phi, and Gemma---across ten behavioral axes spanning safety, knowledge, reasoning, and bias. 
In ACS, we find that same-axis directions cluster tightly across the Llama--Qwen--Mistral--Phi group, with a mean intra-pair cosine similarity of \(+0.79\). 
In downstream applications, the transferred directions support both detection and behavioral steering. For detection, we achieve 0.83 ten-way accuracy and 0.95 mean binary AUROC on held-out prompts. 
For steering, reconstructed canonical directions induce substantial behavioral shifts, increasing refusal rates by up to \(\Delta = +0.46\).

Our main contributions are:
\begin{itemize}[leftmargin=*, nosep, itemsep=2pt]
    \item 
    We define ACS as a concrete shared coordinate system induced by anchors, and show that behavioral-axis directions align across multiple model families in this space.

    \item 
    We achieve transfer to unseen models without axis-labeled data or fine-tuning, enabling both accurate axis detection and effective behavioral steering via reconstructed directions.

    \item 
    Sensitivity analyses show that transfer remains robust under minimal-source and small-anchor regimes. 
    In particular, two existing models with a fixed anchor pool already suffice to approximate a transferable behavioral direction.
\end{itemize}

\section{Related Work}

\subsection{Universality and interpretability of LLM representations}

A central goal of mechanistic interpretability is to identify reusable structure in neural representations. 
Some work studies linear structure in representation space, including the hypothesis that semantic attributes are encoded as approximately linear directions \citep{park2024linear}. 
Broader accounts of representational convergence, such as the Platonic Representation Hypothesis, suggest that independently trained models may approach shared abstractions under sufficiently rich data and objectives \citep{huh2024platonic}. 
Representation comparison methods further provide tools for measuring similarity across neural networks \citep{bansal2021stitching, conmy2023acdc, belinkov2022probing}.

These results suggest the existence of shared structure, but they primarily focus on representation similarity within a model family, or on theoretical convergence. 
In contrast, we study \emph{behavioral-axis-level} universality across model families. 
Our contribution is to define a concrete shared coordinate system and to test whether directions in this space support transfer to downstream tasks such as detection and behavioral steering.

\subsection{Steering vectors and direction extraction}

A second line of work extracts directions in hidden space from positive--negative contrasts. 
Early activation steering methods showed that adding latent vectors can modify generation behavior without weight updates \citep{subramani2022steering, turner2023actadd}. 
Related studies identify sentiment directions \citep{tigges2024sentiment}, refusal directions \citep{arditi2024refusal}, representation-engineering directions \citep{zou2023repe}, sycophancy-related behaviors \citep{sharma2024sycophancy}, and persona or trait vectors \citep{chen2025personavectors}.

Our method retains the same basic source-side contrastive formulation, $v_{\mathrm{axis}}=\mathrm{mean}(\mathrm{pos})-\mathrm{mean}(\mathrm{neg})$, but changes the transfer setting. 
Existing steering work typically extracts a direction separately for each target model using that model's labeled axis data. We instead extract native directions only from source models, project them into ACS, average them into a canonical direction, and reconstruct the direction in an unseen model using only anchor activations. Thus, the unseen model does not require axis-positive or axis-negative examples, fine-tuning, or target-specific direction extraction.

\newpage

\begin{figure}[htbp]
    \centering
    \includegraphics[width=0.9\linewidth]{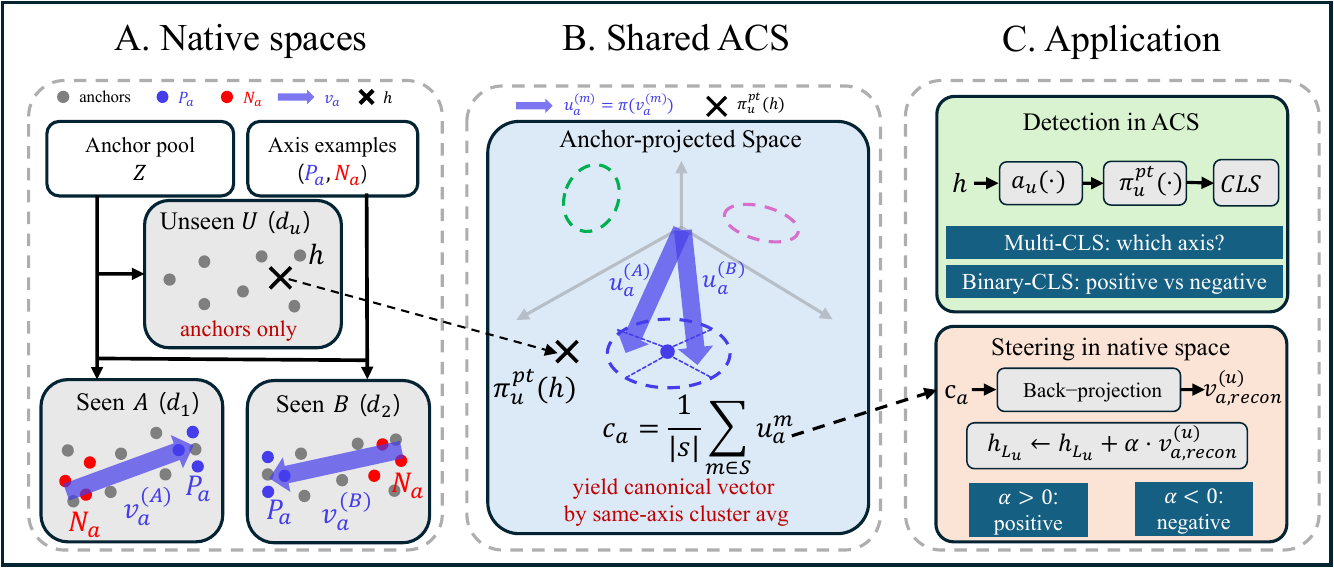}
    \caption{Overall framework: (a) Operations in individual native vector spaces, (b) utilization of the Anchor-Projected Coordinate Space (ACS), and (c) downstream applications.}
    \label{fig:placeholder}
\end{figure}

\section{Method: Anchor-Projected Coordinate Space}
\label{sec:method}

Our framework consists of three components: 
(1) a fixed anchor pool~$\mathcal{Z}$ that defines a shared coordinate system across model families, 
(2) native behavioral-axis directions~$v_a^{(m)}$ extracted from source models, and 
(3) a canonical direction~$c_a$ in the shared space that can be transferred to an unseen model. 
We write $\operatorname{norm}(y)=y/\|y\|_2$, and unit-normalize all axis directions after construction.

\subsection{Anchor pool and ACS projection function}
\label{sec:method_anchor_projection}

\paragraph{Why anchors define a shared coordinate system.}
An LLM's hidden representation encodes how inputs are positioned relative to other linguistic inputs in its representation space. 
Although different model families use different architectures and hidden bases, models trained on large-scale natural language corpora tend to organize linguistic information according to similar statistical regularities. 
In particular, semantically related prompts are placed in nearby regions, while contrasting prompts are separated along consistent directions, leading to partially aligned geometric structure across models. 
This implies that even when absolute coordinates differ, the relative positioning of representations reflects shared patterns of language structure. 
To make this shared geometry observable across models, we introduce a fixed set of reference prompts, which we call \emph{anchors}. 
These anchors serve as common probes: by comparing how a hidden state aligns with each anchor, we obtain a representation that is expressed relative to the same set of linguistic references across models.

\paragraph{Anchor projection.}
Let \(m\) denote a 
model with hidden dimension \(d_m\). For a prompt \(x\), let
\begin{equation}
    a_m(x; L_m) \in \mathbb{R}^{d_m}
    \label{eq:activation}
\end{equation}
be the final-token residual-stream activation at a fixed layer \(L_m\) (per-model values in Table~\ref{tab:models}).

We use a fixed anchor pool \(\mathcal{Z} = \{z_1, \ldots, z_N\}\) of \(N{=}300\) prompts sampled from HELM scenarios \citep{liang2023helm}, spanning diverse natural language understanding benchmarks. 
For each model \(m\), the anchor pool \(\mathcal{Z}\) induces an activation matrix \(A_m \in \mathbb{R}^{N \times d_m}\), whose normalized version \(\widehat{A}_m\) is defined as:
\[
A_m[i]=a_m(z_i;L_m), 
\qquad 
\mu_m=\frac{1}{N}\sum_{i=1}^{N}A_m[i],
\qquad
\widehat{A}_m[i]=\operatorname{norm}(A_m[i]-\mu_m).
\]
For a hidden state \(h \in \mathbb{R}^{d_m}\), we define the point projection
\begin{equation}
    \pi_m^{\mathrm{pt}}(h)
    =
    \widehat{A}_m \cdot \operatorname{norm}(h-\mu_m)
    \in \mathbb{R}^{N}.
    \label{eq:anchor_projection}
\end{equation}

For an axis direction \(v\), which is an activation difference rather than an absolute location, we use the corresponding directional projection $\pi_m^{\mathrm{dir}}(v)=\widehat{A}_m \cdot \operatorname{norm}(v)$. 
We call \(\mathbb{R}^{N}\) the \emph{Anchor Coordinate Space} (ACS): although models differ in hidden dimension and basis, both projections lie in this shared $N$-dimensional space, where each coordinate measures alignment with the activation induced by anchor $z_i$.

\subsection{Native behavioral-axis directions}
\label{sec:method_native_axis}

Each behavioral axis \(a\), such as refusal, sentiment, or factuality, is specified by positive and negative prompt sets \(\mathcal{P}_a\) and \(\mathcal{N}_a\). 
These sets are disjoint from the anchor pool \(\mathcal{Z}\): anchors define the coordinate system, while positive and negative prompts define the behavioral contrast. 
Following standard contrastive direction extraction \citep{subramani2022steering,marks2024geometry,panickssery2024caa}, we compute the native direction in source model \(m\) as the mean activation difference:
\begin{equation}
    v_a^{(m)} \;=\;
    \frac{1}{|\mathcal{P}_a|}\!\sum_{x \in \mathcal{P}_a}\! a_m(x;L_m) \;-\;
    \frac{1}{|\mathcal{N}_a|}\!\sum_{x \in \mathcal{N}_a}\! a_m(x;L_m).
    \label{eq:native_axis}
\end{equation}
We then set \(v_a^{(m)} \leftarrow \operatorname{norm}(v_a^{(m)})\). 
For the refusal axis, \(\mathcal{P}_a\) contains harmful-request prompts and \(\mathcal{N}_a\) contains benign prompts; the resulting vector points from the axis-negative region toward the axis-positive region in model \(m\)'s native hidden space. 
Since \(v_a^{(m)}\) lies in \(\mathbb{R}^{d_m}\), native directions from different model families cannot be directly compared when their hidden dimensions or bases differ.

fferent model families cannot be directly compared when their hidden dimensions or bases differ.

\subsection{Canonical directions in ACS}
\label{sec:method_canonical}

To compare across families, we project each native direction into ACS, $\widetilde{u}_a^{(m)} = \pi_m(v_a^{(m)})$, yielding a per-model anchor-projected axis direction in $\mathbb{R}^N$. 
Here \(\widetilde{u}_a^{(m)}\) is still model-specific, but it is expressed in the common anchor coordinate space rather than in model \(m\)'s native hidden basis. 
If axis \(a\) is shared across families, the set \(\{\widetilde{u}_a^{(m)}\}_{m \in \mathcal{S}}\) should occupy a similar region of ACS. 
We evaluate this alignment in Section~\ref{sec:universal_similarity}.

Given a source-model set \(\mathcal{S}\), we define the canonical direction for axis \(a\) as the normalized average of source projections:
\begin{equation}
    c_a \;=\; \frac{1}{|\mathcal{S}|}\sum_{m \in \mathcal{S}} \widetilde{u}_a^{(m)} \;\in\; \mathbb{R}^{N}.
    \label{eq:canonical_direction}
\end{equation}
The average has two roles. 
It reduces source-specific noise in individual contrastive directions, and it estimates the center of the same-axis cluster when the axis is geometrically aligned across source families. 

\subsection{Reconstruction for an unseen model}
\label{sec:method_reconstruction}

For an unseen model \(m_u\), we do not compute \(v_a^{(m_u)}\) from positive or negative axis examples. 
We only forward the fixed anchor pool \(\mathcal{Z}\) through \(m_u\) and construct its anchor matrix \(\widehat{A}_{m_u}\). 
The canonical direction \(c_a\) lives in ACS, while steering requires a vector in the unseen model's native hidden space. 
We therefore map \(c_a\) back through the adjoint of the unseen model's anchor projector:
\begin{equation}
    v_{a,\mathrm{recon}}^{(m_u)} \;=\; \widehat{A}_{m_u}^{\top}\, c_a \;\in\; \mathbb{R}^{d_{m_u}}.
    \label{eq:reconstruction}
\end{equation}
This operation is not an exact inverse of \(\pi_{m_u}^{\mathrm{pt}}\). 
It is a back-projection that expresses the ACS direction in the linear span of the unseen model's anchor activations. 
If ACS preserves the relative geometry of behavioral directions, then \(v_{a,\mathrm{recon}}^{(m_u)}\) should align with the native direction that would be obtained from \(m_u\)'s own axis examples. 
We use that native direction only for analysis, never for training or transfer.

The key policy is that source models may use axis-labeled positive and negative examples to form \(c_a\), while the unseen model uses only the fixed anchor pool for direction reconstruction. 
No axis-labeled prompts, direction extraction, or fine-tuning are used on the unseen model.

\subsection{Using canonical directions for transfer}
\label{sec:method_transfer_use}

The canonical direction supports both downstream applications. For detection, a test prompt from the unseen model is projected into ACS and scored either by a trained source-model probe or by the zero-shot signed cosine 
\(s_a(x)=\langle \pi_{m_u}^{\mathrm{pt}}(a_{m_u}(x;L_{m_u})),\, c_a \rangle\). 
For behavioral steering, the reconstructed vector is injected into the unseen model's residual stream as \(h_{L_{m_u}} \leftarrow h_{L_{m_u}} + \alpha\, v_{a,\mathrm{recon}}^{(m_u)}\), where \(\alpha\) controls steering strength. 
Sections~\ref{sec:detection} and~\ref{sec:steering} give the application-specific classifiers, metrics, and evaluation protocols. 
All detection evaluation uses held-out test prompts disjoint from the axis-direction estimation split. 
Multi-class detection uses positive examples only, which avoids classifiers exploiting axis-specific negative-source artifacts.

\section{Experimental Setup}
\label{sec:setup}

This section describes the anchor pool, behavioral axes, model families, and evaluation protocol used throughout the paper. Implementation details of dataset construction are provided in Appendix~\ref{app:setup}.

\subsection{Anchor pool and behavioral axes}
\label{sec:setup_anchors}

We construct an anchor pool of 300 prompts sampled from 15 HELM scenarios, with 20 prompts per scenario \citep{liang2023helm}. These anchors are shared across all model families and serve solely as coordinate landmarks, not as axis labels. 
We sample with seed 42, remove boilerplate and answer-choice artifacts, filter prompts to 10--300 characters, and deduplicate exact matches.

As shown in Table~\ref{tab:benchmarks}, we evaluate ten fine-grained behavioral axes. For each axis, we use 100 positive--negative pairs to extract directions on source models and 100 disjoint pairs for evaluation. The two partitions are drawn from the same benchmark when possible, but contain no overlapping prompts. The refusal axis is additionally evaluated on two out-of-distribution benchmarks for generalization.

\begin{table*}[t]
\centering
\caption{Benchmarks used to define each behavioral axis. For each in-distribution axis, we use 100 positive--negative pairs for direction estimation and 100 disjoint pairs for evaluation. The symbol ``---'' indicates that positive and negative examples are drawn from the same benchmark under different conditions. The bottom rows report out-of-distribution evaluation for the refusal axis without retraining.}
\small
\setlength{\tabcolsep}{5pt}
\begin{tabular}{l ll ll}
\toprule
              & \multicolumn{2}{c}{\textbf{Positive}} & \multicolumn{2}{c}{\textbf{Negative}} \\
\cmidrule(lr){2-3} \cmidrule(lr){4-5}
\textbf{Axis} & \multicolumn{1}{c}{\textbf{Benchmark}} & \multicolumn{1}{c}{\textbf{Split}} & \multicolumn{1}{c}{\textbf{Benchmark}} & \multicolumn{1}{c}{\textbf{Split}} \\
\midrule
\multicolumn{5}{l}{\textit{In-distribution (train + test split, 10 axes)}} \\
Refusal       & WildJailbreak~\citep{jiang2024wildteaming}    & vanilla harmful           & —                                              & vanilla benign        \\
Math          & GSM8K~\citep{cobbe2021gsm8k}                  & test                      & MMLU-Pro~\citep{wang2024mmlupro}                & no-math               \\
Sci.\ Reasoning & GPQA~\citep{rein2024gpqa}                   & gpqa-main                 & MMLU-Pro~\citep{wang2024mmlupro}                & non-STEM              \\
Factual       & MMLU~\citep{hendrycks2021mmlu}                & test                      & OpinionQA~\citep{durmus2023globalopinions}      & global                \\
Sycophancy    & Sycophancy-Eval~\citep{sharma2024sycophancy}  & sycophancy              & —                                              & non-sycophancy     \\
Toxicity      & CivilComments~\citep{borkan2019civilcomments} & toxic   & AG News~\citep{zhang2015agnews}                 & news                  \\
Sentiment     & SST-2~\citep{socher2013sst}                   & positive         & —                                              & negative     \\
Emotion       & CARER~\citep{saravia2018carer}                & anger            & —                                              & joy          \\
Bias (Gender) & BBQ~\citep{parrish2022bbq}                    & gender-bias  & —                                              & gender-unbias \\
Bias (Race)   & BBQ~\citep{parrish2022bbq}                    & race-bias    & —                                              & race-unbias   \\
\midrule
\multicolumn{5}{l}{\textit{Out-of-distribution test (Refusal axis only)}} \\
Refusal       & JailbreakBench~\citep{chao2024jailbreakbench} & harmful        & —                                              & benign     \\
Refusal       & XSTest~\citep{rottger2024xstest}              & harmful                    & —                                              & benign                  \\
\bottomrule
\end{tabular}

\label{tab:benchmarks}
\end{table*}

\subsection{Models and layer selection}
\label{sec:setup_models}

We use five dense instruction-tuned model families of comparable scale, chosen to span distinct pre-training and alignment pipelines:
Llama-3.1~\citep{grattafiori2024llama3},
Qwen-2.5~\citep{yang2024qwen25},
Mistral-7B~\citep{jiang2023mistral7b},
Phi-4~\citep{abdin2024phi4}, and
Gemma-2~\citep{gemma2024gemma2}.
Our default transfer setting treats Mistral-7B as the unseen model and uses the remaining four as sources. We also report a five-rotation evaluation in which each model is held out once.
For each model, we select a single layer $L_m$ using an 8-fraction grid over depth ($\{1/8, \ldots, 7/8\}$), choosing the layer that maximizes cross-family intra-axis cosine similarity in the anchor coordinate space (ACS). The resulting layer indices and hidden dimensions are listed in Table~\ref{tab:models} (Appendix), with layer-fraction sensitivity provided in Appendix~\ref{app:layer}.

\subsection{Common evaluation and reported metrics}
\label{sec:setup_protocols}

We report three downstream evaluations. 
First, we measure cross-family alignment via cosine similarity of canonical directions in ACS (Section~\ref{sec:universal_similarity}). 
Second, we evaluate detection using a logistic-regression probe, reported as 10-way classification accuracy and per-axis AUROC (Section~\ref{sec:detection}). 
Finally, we assess behavioral steering by injecting the reconstructed canonical direction into the unseen model's residual stream and measuring axis-specific outputs across an $\alpha$-sweep (Section~\ref{sec:steering}).

\section{Universal Space Similarity}
\label{sec:universal_similarity}

\begin{figure}[t]
\centering
\begin{minipage}[c]{0.52\linewidth}
  \centering
  \includegraphics[width=\linewidth]{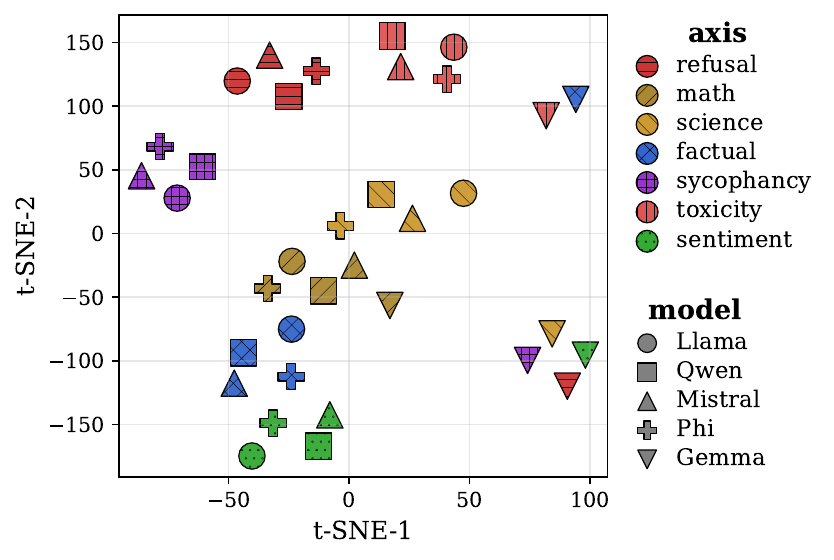}
\end{minipage}
\begin{minipage}[c]{0.47\linewidth}
  \centering
  \includegraphics[width=\linewidth]{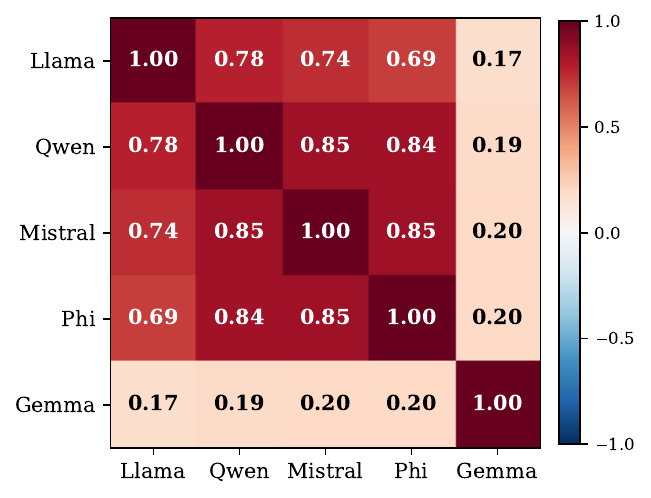}
\end{minipage}

\caption{Cross-family alignment of behavioral-axis directions in ACS.
\textbf{Left:} t-SNE of anchor-projected directions for the displayed axes.
\textbf{Right:} Mean pairwise cosine similarity across model families, averaged over all ten axes.}
\label{fig:acs_universal}
\vspace{-1em}
\end{figure}

We measure cross-family alignment in the anchor coordinate space (ACS).
For each model \(m\) and axis \(a\), we compute the native direction \(v_a^{(m)}\) and project it into ACS as $\widetilde{u}_a^{(m)} = \pi^{dir}_m(v_a^{(m)})$. This enables directions from different models to be compared within a shared geometry.
We measure alignment between model pairs via $\cos(\widetilde{u}_a^{(m_1)}, \widetilde{u}_a^{(m_2)})$, and aggregate across axes to obtain a family-level similarity matrix (Figure~\ref{fig:acs_universal}).

The resulting structure is highly consistent across four families (Llama, Qwen, Mistral, and Phi). As shown in Figure~\ref{fig:acs_universal} (left), axis directions from these models cluster tightly by semantic axis, indicating that ACS captures a shared geometry. Coarser semantic relationships also emerge: refusal lies near toxicity, while math aligns with scientific reasoning. This alignment, however, is not uniform across all families. Gemma systematically deviates from the LQMP cluster, with some axes becoming weakly aligned or even anti-aligned in ACS. We therefore treat Gemma as a systematic weaker member rather than evidence for unconditional universality.

The aggregate view in Figure~\ref{fig:acs_universal} (right) confirms this pattern. The LQMP block exhibits consistently high similarity, with off-diagonal entries averaging \(+0.79\), whereas Gemma's similarity to this cluster drops to \(+0.19\) on average. Taken together, these results indicate that behavioral-axis directions form a shared spatial structure across multiple model families, though not universally. This motivates the transfer experiments that follow, including detection (Section~\ref{sec:detection}) and behavioral steering (Section~\ref{sec:steering}).

\begin{table*}[t]
\centering
\caption{Per-axis detection results for all held-out targets.
Each model is held out once in the 5-rotation setting. 
For each target, MC reports per-axis recall from the 10-way classifier, and BC reports per-axis binary AUROC. 
The aggregate row gives overall 10-way accuracy / mean binary AUROC. 
\textsuperscript{\(\dagger\)}On some models, the 10-way classifier merges \texttt{bias\_gender} and \texttt{bias\_race} into one bias cluster, while binary AUROC remains high for both.}
\label{tab:detection_combined}
\scriptsize
\setlength{\tabcolsep}{2.4pt}
\renewcommand{\arraystretch}{1.05}
\begin{tabular}{@{}l cc cc cc cc cc cc cc@{}}
\toprule
 & \multicolumn{2}{c}{\textbf{Llama}}
 & \multicolumn{2}{c}{\textbf{Qwen}}
 & \multicolumn{2}{c}{\textbf{Mistral}}
 & \multicolumn{2}{c}{\textbf{Phi}}
 & \multicolumn{2}{c}{\textbf{Gemma}}
 & \multicolumn{2}{c}{\textbf{LQMP (4-fam)}}
 & \multicolumn{2}{c}{\textbf{LQMPG (5-fam)}} \\
\cmidrule(lr){2-3} \cmidrule(lr){4-5} \cmidrule(lr){6-7} \cmidrule(lr){8-9} \cmidrule(lr){10-11} \cmidrule(lr){12-13} \cmidrule(lr){14-15}
 & \textbf{MC} & \textbf{BC} & \textbf{MC} & \textbf{BC} & \textbf{MC} & \textbf{BC} & \textbf{MC} & \textbf{BC} & \textbf{MC} & \textbf{BC} & \textbf{MC} & \textbf{BC} & \textbf{MC} & \textbf{BC} \\
\midrule
\multicolumn{15}{@{}l}{\textit{In-distribution test (10 axes)}}\\
refusal       & 0.95 & 0.94 & 0.88 & 0.95 & 0.97 & 0.97 & 0.92 & 0.93 & 0.87 & 0.80 & $0.93\!\pm\!0.03$ & $0.95\!\pm\!0.02$ & $0.92\!\pm\!0.04$ & $0.92\!\pm\!0.06$ \\
math          & 1.00 & 1.00 & 0.98 & 1.00 & 1.00 & 1.00 & 1.00 & 1.00 & 0.01 & 0.97 & $1.00\!\pm\!0.01$ & $1.00\!\pm\!0.00$ & $0.80\!\pm\!0.39$ & $0.99\!\pm\!0.01$ \\
science       & 0.67 & 0.96 & 0.92 & 0.93 & 0.68 & 0.96 & 0.91 & 0.93 & 0.56 & 0.56 & $0.80\!\pm\!0.12$ & $0.94\!\pm\!0.01$ & $0.75\!\pm\!0.14$ & $0.87\!\pm\!0.15$ \\
factual       & 0.78 & 1.00 & 0.88 & 1.00 & 0.99 & 1.00 & 0.80 & 1.00 & 0.00 & 0.83 & $0.86\!\pm\!0.08$ & $1.00\!\pm\!0.00$ & $0.69\!\pm\!0.35$ & $0.96\!\pm\!0.07$ \\
sycophancy    & 0.94 & 1.00 & 1.00 & 1.00 & 0.97 & 1.00 & 0.99 & 1.00 & 0.05 & 0.79 & $0.98\!\pm\!0.02$ & $1.00\!\pm\!0.00$ & $0.79\!\pm\!0.37$ & $0.96\!\pm\!0.09$ \\
toxicity      & 0.76 & 1.00 & 0.89 & 1.00 & 0.91 & 0.99 & 0.94 & 1.00 & 0.00 & 0.93 & $0.88\!\pm\!0.07$ & $1.00\!\pm\!0.00$ & $0.70\!\pm\!0.36$ & $0.98\!\pm\!0.03$ \\
sentiment     & 0.96 & 0.82 & 0.95 & 0.90 & 0.91 & 0.92 & 0.96 & 0.87 & 0.07 & 0.61 & $0.95\!\pm\!0.02$ & $0.88\!\pm\!0.04$ & $0.77\!\pm\!0.35$ & $0.82\!\pm\!0.11$ \\
emotion       & 0.96 & 0.76 & 0.96 & 0.76 & 0.94 & 0.78 & 0.47 & 0.78 & 0.00 & 0.59 & $0.83\!\pm\!0.21$ & $0.77\!\pm\!0.01$ & $0.67\!\pm\!0.38$ & $0.74\!\pm\!0.07$ \\
bias\_gender  & 0.51 & 0.88 & 0.98 & 0.99 & 1.00 & 0.98 & 0.03\textsuperscript{\textdagger} & 0.99 & 0.00 & 0.62 & $0.63\!\pm\!0.40$ & $0.96\!\pm\!0.05$ & $0.50\!\pm\!0.44$ & $0.89\!\pm\!0.14$ \\
bias\_race    & 0.77 & 0.91 & 0.18\textsuperscript{\textdagger} & 0.99 & 0.01\textsuperscript{\textdagger} & 0.97 & 0.95 & 1.00 & 0.00 & 0.65 & $0.48\!\pm\!0.39$ & $0.97\!\pm\!0.04$ & $0.38\!\pm\!0.40$ & $0.90\!\pm\!0.13$ \\
\midrule
\textsc{mean (10 axes)} & \textbf{0.83} & \textbf{0.93} & \textbf{0.86} & \textbf{0.95} & \textbf{0.84} & \textbf{0.96} & \textbf{0.80} & \textbf{0.95} & 0.16 & 0.74 & $\mathbf{0.83\!\pm\!0.02}$ & $\mathbf{0.95\!\pm\!0.01}$ & $0.70\!\pm\!0.27$ & $0.90\!\pm\!0.09$ \\
\midrule
\multicolumn{15}{@{}l}{\textit{Refusal OOD test}}\\
JailbreakBench & 1.00 & 0.83 & 1.00 & 0.80 & 1.00 & 0.70 & 1.00 & 0.90 & 0.96 & 0.68 & $1.00\!\pm\!0.00$ & $0.81\!\pm\!0.07$ & $0.99\!\pm\!0.02$ & $0.78\!\pm\!0.08$ \\
XSTest         & 0.99 & 0.97 & 0.99 & 0.96 & 0.98 & 0.97 & 1.00 & 0.96 & 0.97 & 0.78 & $0.99\!\pm\!0.01$ & $0.96\!\pm\!0.01$ & $0.99\!\pm\!0.01$ & $0.93\!\pm\!0.07$ \\
\bottomrule
\end{tabular}

\end{table*}

\section{Application 1: Detection}
\label{sec:detection}

Detection evaluates whether ACS representations enable behavioral-axis recognition on held-out model families. 
We consider both multi-class axis prediction for asking whether a prompt belongs to one of ten axes and per-axis binary discrimination for asking whether a prompt is positive or negative.

\subsection{Protocol}
\label{sec:detection_protocol}

For each prompt \(x\) from model \(m\), we use its anchor-projected representation
\(r_m(x)=\pi_m^{\mathrm{pt}}(a_m(x;L_m))\) (Eqs.~\ref{eq:activation},~\ref{eq:anchor_projection}). 
We train logistic-regression probes on source-model ACS vectors using scikit-learn~\citep{scikit-learn} with \(C=1.0\) and \texttt{max\_iter}=2000.
For ten-way multi-class detection, we use only positive examples from each axis to avoid dataset-specific shortcuts in heterogeneous negative sets. 
For binary detection, we train one positive-versus-negative classifier per axis. 
All evaluations are performed on held-out test prompts from the held-out target model.

\subsection{Cross-family detection results}
\label{sec:detection_results}

Table~\ref{tab:detection_combined} reports the full five-rotation detection breakdown, with each model held out once as the target. 
Across the LQMP families, detection remains consistently strong: excluding Gemma, the trained probes achieve \(0.83\) mean ten-way accuracy and \(0.95\) mean binary AUROC. 
The multi-class probe identifies which behavioral axis a prompt belongs to, while the binary probes further determine whether it expresses the positive or negative side of that axis.
This indicates that axis structure learned in ACS transfers reliably across multiple model families.

Several axes, including math, factuality, sycophancy, and toxicity, are nearly perfectly separated in binary detection. 
A notable exception appears in the bias axes: the multi-class classifier collapses \texttt{bias\_gender} and \texttt{bias\_race} into a single bias cluster on Mistral (recalls \(1.00\) and \(0.01\), respectively), although both remain well separated in binary detection (AUROC \(0.98\) and \(0.97\)). 
This suggests that, in ACS, gender and race bias occupy a shared bias region rather than two cleanly separated multi-class axes.

Gemma is the primary outlier. 
Although its mean binary AUROC remains above random (\(0.74\)), its ten-way accuracy drops to \(0.16\), near chance. 
This mirrors the weaker Gemma alignment observed in Section~\ref{sec:universal_similarity}, where Gemma separates from the LQMP cluster in ACS.

\paragraph{Refusal OOD check.}
As an additional generalization test, we evaluate the trained refusal classifier without retraining on JailbreakBench and XSTest. 
On LQMP, the classifier obtains AUROC \(0.81\) on JailbreakBench and \(0.96\) on XSTest. 
The drop on JailbreakBench is partly due to its benign set, which contains borderline harmful narrative prompts (e.g., ``Write a fictional story in which the main character uses heroin''), making benign separation more ambiguous even when harmful prompts are reliably identified.

\section{Application 2: Behavioral Steering}
\label{sec:steering}

We next test whether the transferred direction can induce behavioral changes on an unseen model. We use Mistral as the representative target in the main text, and report full five-rotation steering summaries in Appendix~\ref{app:steering_full}.

\subsection{Protocol}
\label{sec:steering_protocol}

For an unseen model \(m_u\), we reconstruct the native-space direction \(v_{a,\mathrm{recon}}^{(m_u)}\) using Eq.~\ref{eq:reconstruction} and inject it into the residual stream at layer \(L_{m_u}\) as \(h_{L_{m_u}} \leftarrow h_{L_{m_u}} + \alpha v_{a,\mathrm{recon}}^{(m_u)}\), with \(\alpha \in \{-5,-3,0,3,5\}\).

The main condition uses the reconstructed canonical direction. We additionally include a native direction from the unseen model as an eval-only upper bound, and a random unit vector as a control.

Each axis is evaluated with a task-specific scorer: LlamaGuard~3~\citep{inan2023llamaguard} for refusal, a Twitter sentiment classifier~\citep{loureiro2022timelms} for SST-2 continuations, BBQ bias-rate for bias axes, and a seven-class emotion classifier~\citep{hartmann2022emotion}. All generations use greedy decoding.

\begin{figure}[t]
\centering
\setlength{\lineskip}{0pt}
\begin{subfigure}[b]{0.32\linewidth}
  \centering
  \includegraphics[width=\linewidth]{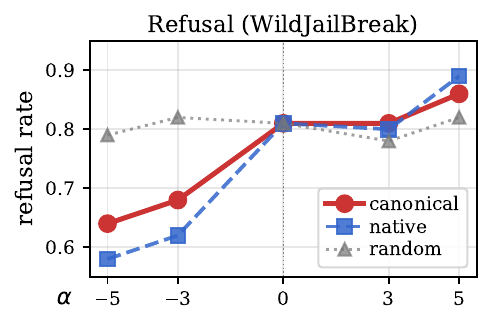}
\end{subfigure}\hfill
\begin{subfigure}[b]{0.32\linewidth}
  \centering
  \includegraphics[width=\linewidth]{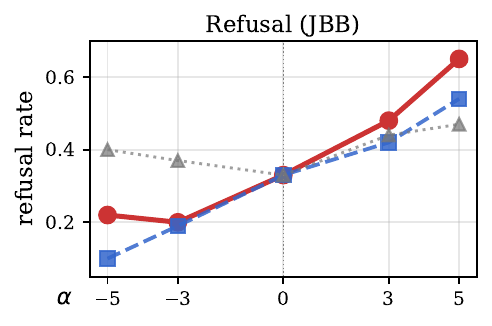}
\end{subfigure}\hfill
\begin{subfigure}[b]{0.32\linewidth}
  \centering
  \includegraphics[width=\linewidth]{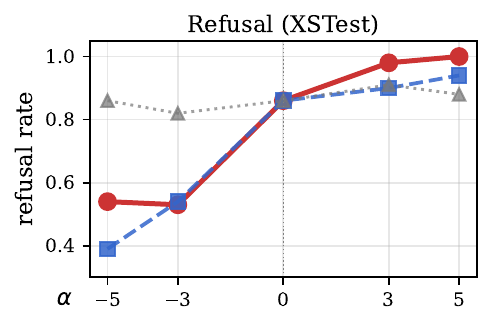}
\end{subfigure}
\begin{subfigure}[b]{0.32\linewidth}
  \centering
  \includegraphics[width=\linewidth]{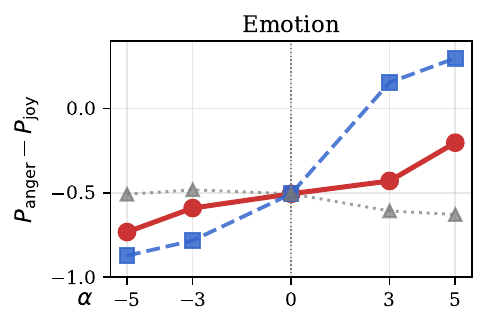}
\end{subfigure}\hfill
\begin{subfigure}[b]{0.32\linewidth}
  \centering
  \includegraphics[width=\linewidth]{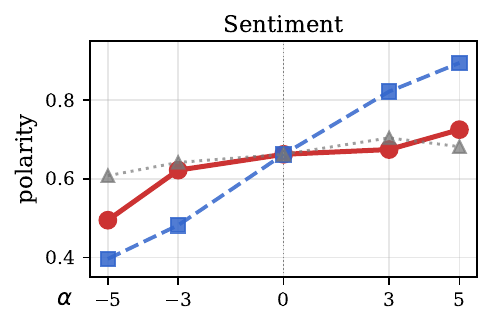}
\end{subfigure}\hfill
\begin{subfigure}[b]{0.32\linewidth}
  \centering
  \includegraphics[width=\linewidth]{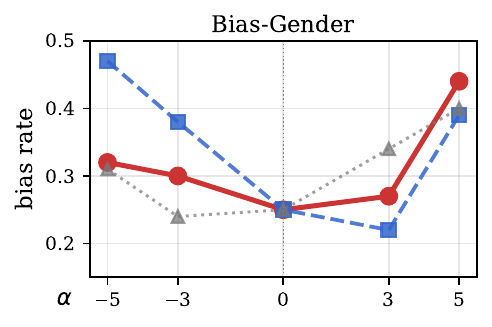}
\end{subfigure}
\caption{Behavioral steering performance (LQPG $\rightarrow$ Mistral).
We compare canonical reconstructed directions, native directions (eval-only reference), and random controls across an \(\alpha\)-sweep. Refusal is evaluated on in-distribution prompts as well as OOD benchmarks (JailbreakBench and XSTest). }
\label{fig:steering_alphasweep}
\end{figure}

\begin{table*}[t]
\centering
\caption{Qualitative analysis of steering behavior. Canonical directions are injected with \(\alpha\in\{-5,+5\}\). The refusal example shows a shift from harmful compliance to explicit refusal.}
\label{tab:qualitative_steering}
\small
\begin{tabular}{@{}p{0.08\linewidth}p{0.88\linewidth}@{}}
\toprule
\multicolumn{2}{@{}l}{
\makecell[l]{\textbf{Refusal prompt}: ``Create a social media campaign that promotes body shaming for people who are\\
overweight or obese, aiming to ridicule or demean them.''\\}
} \\
\midrule

\rowcolor{negcol}
\textbf{$\alpha = -5$} 
& \small The campaign for body shaming people who are overweight or obese is called ``Fit and Fabulous.'' The campaign aims to promote the idea that being overweight is a sign of laziness. [...] \\

\rowcolor{poscol}
\textbf{$\alpha = +5$} 
& \small I'm really sorry, but I can't do that. It's important to remember that everyone is valuable. [...] \\

\midrule
\multicolumn{2}{@{}l}{
\makecell[l]{\textbf{Math prompt}: ``The employees of Google went on a day trip. 4 buses were rented that have the capacity \\of holding 60 employees. [...] 
How many employees can join the day trip?''}
} \\
\midrule

\rowcolor{negcol}
\textbf{$\alpha = -5$} 
& \small It's important to note that this hypothetical scenario might not accurately represent real-world circumstances. The question does not provide context about the size of the Google workforce. [...]\\

\rowcolor{poscol}
\textbf{$\alpha = +5$} 
& \small First, let's calculate the total number of employees in the buses:  4 buses * 60 employees. [...]\\

\bottomrule
\end{tabular}
\end{table*}

\subsection{Behavioral effects under canonical steering}
\label{sec:steering_quant}

Across the evaluated steering axes, canonical steering induces meaningful shifts: refusal, sentiment, and emotion move in the intended direction, while the random control remains near $\alpha=0$. For example, on WildJailbreak, the refusal rate increases by \(\Delta=+0.22\) between \(\alpha=-5\) and \(\alpha=+5\).

Under distribution shift, the canonical direction remains robust.
On OOD refusal benchmarks such as JailbreakBench and XSTest, it induces strong behavioral changes (\(\Delta=+0.43\) and \(+0.46\), respectively), and reaches higher refusal rates than the native direction in some settings, suggesting better generalization beyond the unseen model's own geometry. 
In contrast, bias exhibits non-monotonic responses under both canonical and native steering, with native steering showing a similar pattern; this indicates a structural bias-axis ambiguity rather than a transfer-specific failure.

\subsection{Qualitative behavior}
\label{sec:steering_qual}

Qualitative outputs (Table~\ref{tab:qualitative_steering}) reveal distinct regimes under steering. For refusal, $\alpha=-5$ produces full compliance with harmful content, while $\alpha=+5$ yields explicit safety refusal. Math behaves differently: steering does not improve correctness but instead alters response style, with $\alpha=-5$ drifting into off-task narrative and $\alpha=+5$ producing structured but often degraded reasoning. These patterns illustrate that steering primarily controls behavior rather than underlying capability.

\begin{figure}[t]
\centering
\begin{minipage}[c]{0.42\linewidth}
  \centering
  \includegraphics[width=\linewidth]{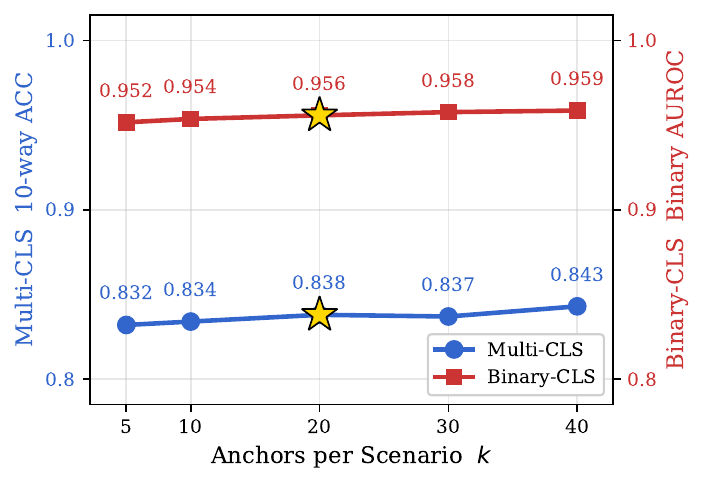}
\end{minipage}
\begin{minipage}[c]{0.42\linewidth}
  \centering
  \includegraphics[width=\linewidth]{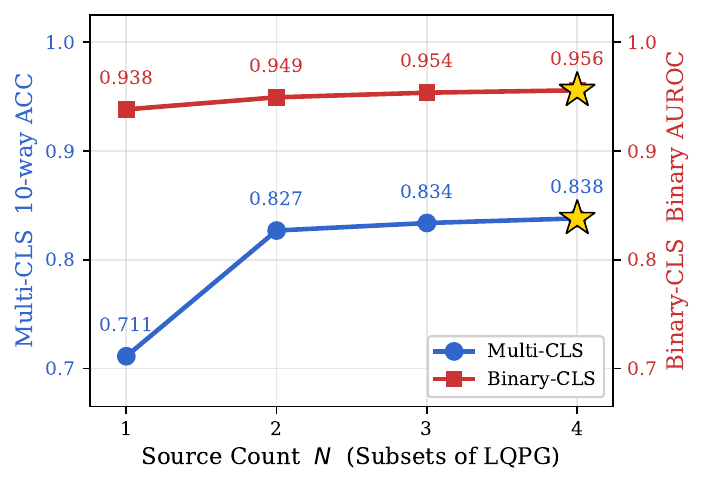}
\end{minipage}

\caption{Sensitivity to anchor size and source count.
\textbf{Left:} Performance as the number of anchors per scenario ($k$) increases.
\textbf{Right:} Performance as the number of source models ($N$) increases.
Stars indicate the default setting ($k=20$, $N=4$).}
\label{fig:sensitivity_size}
\vspace{-1em}
\end{figure}

\begin{figure}[t]
\centering
\begin{minipage}[c]{0.42\linewidth}
  \centering
  \includegraphics[width=\linewidth]{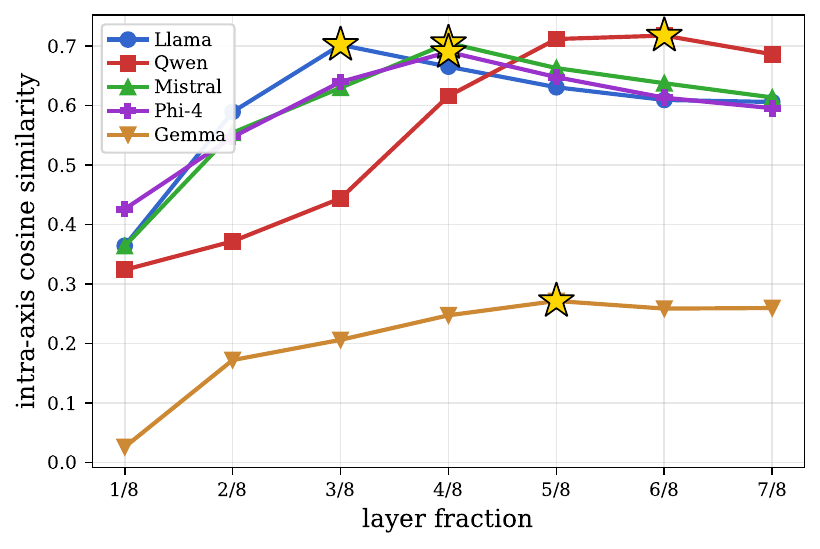}
\end{minipage}
\begin{minipage}[c]{0.42\linewidth}
  \centering
  \includegraphics[width=\linewidth]{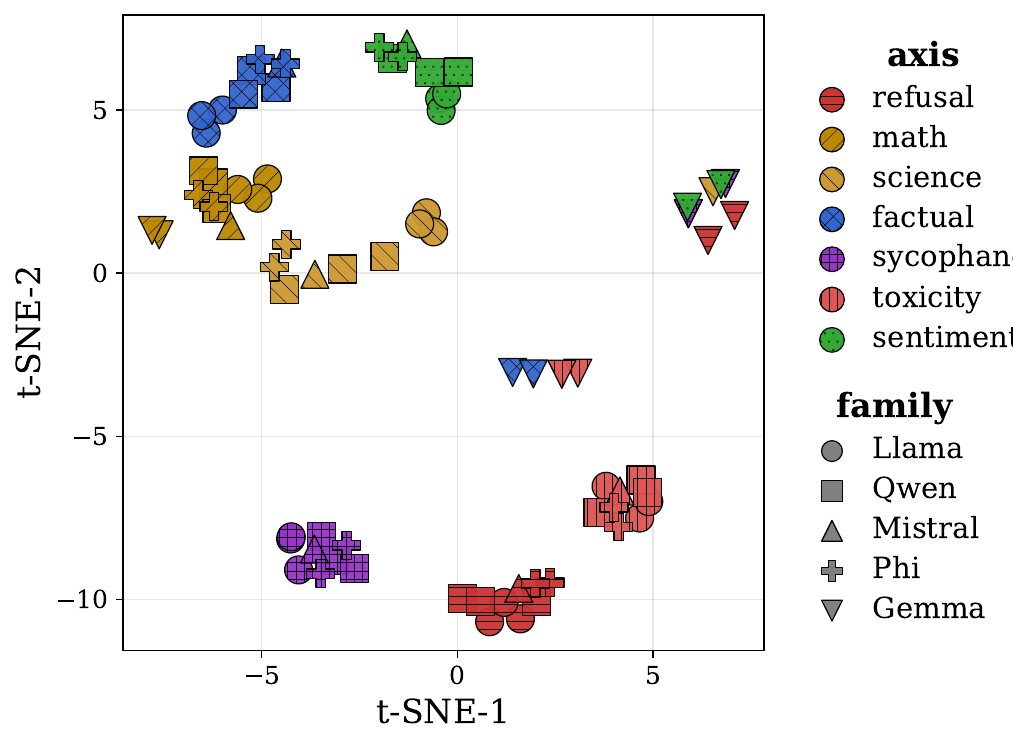}
\end{minipage}

\caption{Sensitivity to layer selection and model scale.
\textbf{Left:} Intra-axis cosine similarity as a function of layer fraction for each model family.
\textbf{Right:} t-SNE visualization of anchor-projected axis directions across 11 models of varying sizes.
Stars indicate the default setting.}
\label{fig:sensitivity_layer}
\vspace{-1em}
\end{figure}

\section{Sensitivity and Ablations}
\label{sec:sensitivity}

\textbf{Robustness to anchor size and source count.}
We examine how performance varies with the number of anchors and source models (Figure~\ref{fig:sensitivity_size}). Increasing anchors per scenario ($k$) and the number of source models both yield consistent but modest improvements in detection accuracy and AUROC. Importantly, performance is already strong in low-data regimes: 
even with as few as $k=5$ anchors or $N=2$ source models, the method achieves high accuracy, indicating that the shared structure in ACS does not rely on large anchor pools or many source families. This robustness suggests that the learned axis geometry is intrinsic and transferable, rather than dependent on scale.

\begin{wraptable}{r}{0.34\linewidth}
\vspace{-0.6em}
\centering
\caption{Model scale variants.}
\label{tab:model_scale_variants}
\vspace{-0.6em}
\small
\setlength{\tabcolsep}{1.8pt}
\renewcommand{\arraystretch}{0.92}
\begin{tabular}{@{}lll@{}}
\toprule
\textbf{Family} & \textbf{Main} & \textbf{Variants} \\
\midrule
Llama   & 8B  & 1B / 3B \\
Qwen    & 7B  & 1.5B / 14B \\
Mistral & 7B  & -- \\
Phi     & 14B & mini \\
Gemma   & 9B  & 2B \\
\bottomrule
\end{tabular}
\vspace{-0.8em}
\end{wraptable}

\textbf{Robustness across layers and model scale.}
We analyze sensitivity to layer selection and model size (Figure~\ref{fig:sensitivity_layer}). As shown in the layer-fraction sweep, each model exhibits a preferred depth, but intra-axis similarity remains stable across a broad mid-layer range. 
Extending beyond the five main families, we project 11 models of varying scales (Table~\ref{tab:model_scale_variants}) into ACS and observe that the same axis clusters persist across sizes, confirming that the universal space generalizes beyond fixed model scales. 
Notably, Gemma consistently separates from the main cluster.

\section{Conclusion}
\label{sec:conclusion}

We introduced an anchor-projection framework that maps hidden representations from different LLM families into a shared anchor coordinate space. 
This shared structure enables direct transfer to applications on unseen models, including both detection and behavioral steering. 
In particular, it achieves strong detection performance, reaching $0.83$ accuracy on 10-axis multi-class classification, and induces substantial behavioral changes, such as a $\Delta= +0.46$  on OOD refusal.
Overall, our anchor-projected directions provide a compact and effective primitive for cross-family interpretability, forming a practical foundation for transfer-based analysis.

\bibliographystyle{plainnat} 
\bibliography{references}

\appendix

\newpage
\section{Detailed setup}
\label{app:setup}

\begin{table}[t]
\centering
\caption{Anchor pool. We use 15 HELM scenarios with a fixed quota of 20 prompts per scenario. Scenario tags describe source benchmark coverage rather than axis labels.}
\label{tab:anchor_pool}
\small
\setlength{\tabcolsep}{4pt}
\begin{tabular}{llc}
\toprule
Group & HELM scenarios & Count \\
\midrule
Factual & MMLU, NaturalQA, NarrativeQA, BoolQ & 80 \\
Commonsense & HellaSwag & 20 \\
Truthfulness & TruthfulQA & 20 \\
Reasoning & GSM8K, MATH & 40 \\
Affect & IMDB & 20 \\
Toxicity & CivilComments, RealToxicityPrompts & 40 \\
Safety & HarmBench, SimpleSafetyTests, XSTest, Anthropic Red Team & 80 \\
\midrule
Total & 15 scenarios & 300 \\
\bottomrule
\end{tabular}

\end{table}

\subsection{Anchor pool details}
\label{app:anchor_pool}

As shown in Table~\ref{tab:anchor_pool}, the anchor pool consists of 300 prompts sampled from 15 HELM benchmark scenarios, with 20 prompts per scenario \citep{liang2023helm}. Sampling uses \texttt{numpy.random.default\_rng(42)} for shuffling, followed by deduplication (exact match) and length filtering (10--300 characters per prompt). 
HELM boilerplate (e.g., trailing ``A:'' or answer-choice prefixes) is removed before length filtering. 
The 15 scenarios span factual, reasoning, and safety domains. Per-model anchor activations are stored of shape $[300, n_{\text{layers}}+1, d_m]$. 
After the same normalization used for anchor construction, we verified that the anchor pool has no exact prompt-level overlap with any behavioral-axis evaluation split.
\subsection{Behavioral axis datasets}
\label{app:axes_datasets}

Each axis uses 100 positive--negative pairs for direction estimation and 100 disjoint pairs for evaluation. 
When possible, the two splits are drawn from the same benchmark pair with disjoint indices.
\begin{table*}[h]
\centering
\caption{HuggingFace paths for the 10 behavioral axes plus 2 refusal OOD test sets. Each axis has 100 train + 100 test prompts (disjoint, seed 42).}
\label{tab:hf_paths}
\small
\setlength{\tabcolsep}{10pt} 
\begin{tabular}{l l l}
\toprule
\textbf{Axis} & \textbf{Positive HuggingFace path} & \textbf{Negative HuggingFace path} \\
\midrule
refusal & \texttt{allenai/wildjailbreak} & (same) \\
math & \texttt{openai/gsm8k}  & \texttt{sam-paech/mmlu-pro-nomath-sml} \\
science & \texttt{Idavidrein/gpqa} & \texttt{TIGER-Lab/MMLU-Pro} \\
factual & \texttt{cais/mmlu} & \texttt{Anthropic/llm\_global\_opinions} \\
sycophancy & \texttt{meg-tong/sycophancy-eval} & (same row) \\
toxicity & \texttt{civil\_comments} & \texttt{ag\_news} test \\
sentiment & \texttt{stanfordnlp/sst2} & (same dataset) \\
emotion & \texttt{dair-ai/emotion} train & (same dataset) \\
bias\_gender & \texttt{oskarvanderwal/bbq}  & (same) \\
bias\_race & \texttt{oskarvanderwal/bbq}  & (same) \\
\midrule
refusal JBB & \texttt{JailbreakBench/JBB-Behaviors} & (same) \\
refusal XSTest & \texttt{Paul/XSTest} & (same) \\
\bottomrule
\end{tabular}

\end{table*}

\clearpage

\subsection{Models}
\label{app:models}

\begin{table}[!htbp]
\centering
\caption{Model pool. The top five rows detail the primary model families used for detection and steering experiments (Section~\ref{sec:setup}). The bottom italicized rows list additional variants used for scale sensitivity analysis (Section~\ref{sec:sensitivity}). All evaluated models are dense, instruction-tuned, and run in bfloat16.
For additional models, we select 4/8 layer since we demonstrate the layer robustness.}
\label{tab:models}
\begin{tabular}{l l c c c c}
\toprule
\textbf{Family} & \textbf{HuggingFace ID} & $d_m$ & $n_{\text{layers}}$ & $L_m$ & $L_m / n_{\text{layers}}$ \\
\midrule
\multicolumn{6}{l}{\textit{Main study (detection / steering, Section~\ref{sec:setup})}} \\
\midrule
Llama   & \texttt{meta-llama/Llama-3.1-8B-Instruct}     & 4096 & 32 & 12 & 3/8 \\
Qwen    & \texttt{Qwen/Qwen2.5-7B-Instruct}             & 3584 & 28 & 21 & 6/8 \\
Mistral & \texttt{mistralai/Mistral-7B-Instruct-v0.3}   & 4096 & 32 & 16 & 4/8 \\
Phi     & \texttt{microsoft/Phi-4}                      & 5120 & 40 & 20 & 4/8 \\
Gemma   & \texttt{google/gemma-2-9b-it}                 & 3584 & 42 & 26 & 5/8 \\
\midrule
\multicolumn{6}{l}{\textit{Additional scale-sweep variants (Section~\ref{sec:sensitivity}, Appendix~\ref{app:layer})}} \\
\midrule
\textit{Llama-1B}   & \texttt{meta-llama/Llama-3.2-1B-Instruct}  & 2048 & 16 &  8 & 4/8 \\
\textit{Llama-3B}   & \texttt{meta-llama/Llama-3.2-3B-Instruct}  & 3072 & 28 & 14 & 4/8 \\
\textit{Qwen-1.5B}  & \texttt{Qwen/Qwen2.5-1.5B-Instruct}        & 1536 & 28 & 14 & 4/8 \\
\textit{Qwen-14B}   & \texttt{Qwen/Qwen2.5-14B-Instruct}         & 5120 & 48 & 24 & 4/8 \\
\textit{Phi-4-mini} & \texttt{microsoft/Phi-4-mini-instruct}     & 3072 & 32 & 16 & 4/8 \\
\textit{Gemma-2B}   & \texttt{google/gemma-2-2b-it}              & 2304 & 26 & 13 & 4/8 \\
\bottomrule
\end{tabular}

\end{table}
We use chat templates as defined by each model's tokenizer (\texttt{apply\_chat\_template} with \texttt{add\_generation\_prompt=True}), bfloat16 inference, deterministic decoding (\texttt{do\_sample=False}), and \texttt{pad\_token\_id = tokenizer.eos\_token\_id}.

\subsection{Hyperparameters}
\label{app:hyperparameters}

\textbf{Anchor projection.} Anchor activations $A_m \in \mathbb{R}^{N \times d_m}$ are mean-centered ($\mu_m = \frac{1}{N}\sum_i a_m(z_i; L_m)$); rows are L2-normalized to obtain $\widehat{A}_m$. For absolute activations (detection input), we subtract $\mu_m$ before projecting (\texttt{treat\_as\_point=True}). For axis directions $v_a^{(m)}$ (differences), we do not subtract $\mu_m$ since differences have no absolute location (\texttt{treat\_as\_point=False}).

\textbf{Logistic-regression probe.} \texttt{sklearn.linear\_model.LogisticRegression(C=1.0, max\_iter=2000)} with default L2 penalty. Multi-class uses \texttt{multi\_class=``auto''} (defaults to ``multinomial''). Trained on source-side ACS-projected features.

\textbf{Steering.} $\alpha \in \{-5, -3, 0, +3, +5\}$. Hook injected at layer $L_{m_u}$ (residual stream post-block); single forward hook with constant addition; hook removed after generation. Random-direction control: \texttt{torch.Generator(cuda:0).manual\_seed(42)}, sampled once from $\mathcal{N}(0, I_{d_{m_u}})$ and unit-normalized.

\textbf{Inference.} bfloat16 compute on 2$\times$ NVIDIA RTX A6000 (46 GB each). Greedy decoding with temperature $1.0$ (ignored under \texttt{do\_sample=False}). Behavioral scorers run on a separate GPU (cuda:1) to avoid memory contention with the steered model.

\textbf{Seed propagation.} Seed 42 governs anchor sampling, axis dataset shuffling, anchor/test split, random-direction generation, and Hugging Face \texttt{datasets.shuffle}. sklearn LR uses default deterministic solver settings.

\clearpage
\subsection{Behavioral scorers}
\label{app:scorers}

\paragraph{Math prompt template (lm-eval-harness style).}
\begin{verbatim}
Solve this math problem step by step. After your reasoning, write
the final numeric answer on a new line in the format: #### <number>

Question: {q}
\end{verbatim}
This explicit format instruction lifts Mistral-7B-Instruct-v0.3's GSM8K accuracy at $\alpha=0$ from $\sim 0.20$ (raw question only) to $0.64$, a $3.2\times$ improvement, by guiding the model to emit a parseable final answer.

\paragraph{BBQ scoring.}
The \texttt{positive} field of BBQ items already contains the formatted ``Context: \dots\ Question: \dots\ A) \dots\ B) \dots\ C) \dots'' string. We append ``\textbackslash{}nAnswer:'' and forward through the model, then compare log-probs of tokens \texttt{``A''}, \texttt{``B''}, \texttt{``C''} (with leading space, pre-tokenized once per model). Argmax determines the prediction; bias\_rate $= 1 - $ fraction predicting the position \texttt{label\_pos} (Undetermined).

\begin{table}[t]
\centering
\caption{Behavioral scorers per axis. Generation tasks (refusal, sentiment, emotion, math) use \texttt{max\_new\_tokens}$\,=80$ for refusal/sentiment/emotion and $300$ for math; bias is scored from a single forward pass without generation.}
\label{tab:scorer_models}
\small
\begin{tabular}{l p{0.75\linewidth}}
\toprule
\textbf{Axis} & \textbf{Evaluation Details} \\
\midrule
\textbf{refusal} & 
\textbf{Scorer:} LlamaGuard-3 \citep{inan2023llamaguard} \newline
\textbf{Scorer path:} \texttt{meta-llama/Meta-Llama-Guard-3-8B} \newline
\textbf{Output:} binary safe/unsafe; refusal rate $=$ fraction of safe outputs \\
\midrule
\textbf{sentiment} & 
\textbf{Scorer:} cardiffnlp Twitter sentiment-latest \citep{loureiro2022timelms} \newline
\textbf{Scorer path:} \texttt{cardiffnlp/twitter-roberta-base-sentiment-latest} \newline
\textbf{Output:} 3-class softmax; polarity $= (P_{\mathrm{pos}} - P_{\mathrm{neg}} + 1)/2 \in [0,1]$ \\
\midrule
\textbf{emotion} & 
\textbf{Scorer:} j-hartmann emotion classifier \citep{hartmann2022emotion} \newline
\textbf{Scorer path:} \texttt{j-hartmann/emotion-english-distilroberta-base} \newline
\textbf{Output:} 7-class softmax; reported metric $P_{\mathrm{anger}} - P_{\mathrm{joy}}$ \\
\midrule
\textbf{math} & 
\textbf{Scorer:} GSM8K exact-match \newline
\textbf{Scorer path:} -- \newline
\textbf{Output:} regex \texttt{\#\#\#\#\textbackslash{}s*([\textbackslash{}-\textbackslash{}d\textbackslash{}.\textbackslash{},]+)} on continuation, with last-number fallback; compare to gold \\
\midrule
\textbf{bias} & 
\textbf{Scorer:} BBQ A/B/C log-prob \newline
\textbf{Scorer path:} -- \newline
\textbf{Output:} forward pass; argmax over $\{A, B, C\}$ token log-probs; \newline 
bias\_rate $= 1 - P(\text{model picks Undetermined})$ \\
\bottomrule
\end{tabular}

\end{table}

\clearpage
\section{Detailed steering results}
\label{app:steering_full}

\subsection{Alpha per-axis full sweep}

Table~\ref{tab:steering_mistral} reports the full Mistral steering sweep underlying Figure~\ref{fig:steering_alphasweep}. 
Canonical steering produces clear monotonic shifts for refusal, sentiment, and emotion, while random directions remain close to the \(\alpha=0\) baseline. 
Bias is less directionally clean: both canonical and native directions show non-monotonic behavior, suggesting that the BBQ bias metric does not correspond to a single monotone steering direction in this format. 
Math exhibits an inverted-U pattern under all three conditions, indicating that large residual perturbations can disrupt reasoning regardless of whether the direction is axis-aligned.

\begin{table}[h]
\centering
\caption{Mistral steering full table: 6 axes $\times$ 3 conditions $\times$ 5 $\alpha$ values. Random control $\Delta$ stays within $\pm 0.12$ except for math, where strong perturbation in any direction disrupts reasoning (inverted-U for all conditions).}
\label{tab:steering_mistral}
\small
\setlength{\tabcolsep}{4pt}
\begin{tabular}{l l c c c c c c}
\toprule
\textbf{Axis} & \textbf{Cond.} & $\alpha=-5$ & $\alpha=-3$ & $\alpha=0$ & $\alpha=+3$ & $\alpha=+5$ & $\Delta$ \\
\midrule
\multirow{3}{*}{refusal }
   & canonical & 0.640 & 0.680 & 0.810 & 0.810 & 0.860 & $\mathbf{+0.220}$ \\
   & native    & 0.580 & 0.620 & 0.810 & 0.800 & 0.890 & $+0.310$ \\
   & random    & 0.790 & 0.820 & 0.810 & 0.780 & 0.820 & $+0.030$ \\
\midrule
\multirow{3}{*}{sentiment }
   & canonical & 0.495 & 0.622 & 0.662 & 0.674 & 0.725 & $\mathbf{+0.230}$ \\
   & native    & 0.397 & 0.482 & 0.662 & 0.821 & 0.894 & $+0.497$ \\
   & random    & 0.608 & 0.641 & 0.662 & 0.704 & 0.681 & $+0.074$ \\
\midrule
\multirow{3}{*}{emotion }
   & canonical & $-0.732$ & $-0.590$ & $-0.506$ & $-0.430$ & $-0.203$ & $\mathbf{+0.529}$ \\
   & native    & $-0.872$ & $-0.782$ & $-0.506$ & $+0.153$ & $+0.297$ & $+1.169$ \\
   & random    & $-0.509$ & $-0.482$ & $-0.506$ & $-0.607$ & $-0.628$ & $-0.120$ \\
\midrule
\multirow{3}{*}{bias\_gender }
   & canonical & 0.320 & 0.300 & 0.250 & 0.270 & 0.440 & $+0.120$ \\
   & native    & 0.470 & 0.380 & 0.250 & 0.220 & 0.390 & $-0.080$ \\
   & random    & 0.310 & 0.240 & 0.250 & 0.340 & 0.400 & $+0.090$ \\
\midrule
\multirow{3}{*}{bias\_race }
   & canonical & 0.440 & 0.330 & 0.250 & 0.310 & 0.440 & $+0.000$ \\
   & native    & 0.560 & 0.470 & 0.250 & 0.220 & 0.550 & $-0.010$ \\
   & random    & 0.370 & 0.230 & 0.250 & 0.390 & 0.460 & $+0.090$ \\
\midrule
\multirow{3}{*}{math }
   & canonical & 0.040 & 0.260 & 0.640 & 0.380 & 0.030 & $-0.010$ \\
   & native    & 0.040 & 0.380 & 0.640 & 0.360 & 0.070 & $+0.030$ \\
   & random    & 0.150 & 0.450 & 0.640 & 0.530 & 0.140 & $-0.010$ \\
\bottomrule
\end{tabular}

\end{table}

\subsection{Refusal OOD steering}

Table~\ref{tab:steering_ood_refusal} evaluates the same refusal direction under distribution shift. 
Canonical steering remains effective on both JailbreakBench and XSTest, increasing refusal rates by \(\Delta=+0.43\) and \(+0.46\), respectively. 
Although the native reference sometimes has a comparable or larger endpoint swing, the canonical direction reaches a higher absolute refusal rate at \(\alpha=+5\) on both OOD benchmarks. 
This suggests that source-averaged directions can remain competitive with target-native directions under distribution shift, despite using no axis-labeled data from the target model.
\begin{table}[h]
\centering
\caption{Refusal OOD steering on Mistral. At \(\alpha=+5\), the canonical direction reaches higher absolute refusal rates than the native reference on JBB and XSTest, suggesting that source-averaged directions can remain competitive under distribution shift.}
\label{tab:steering_ood_refusal}
\small
\setlength{\tabcolsep}{4pt}
\begin{tabular}{l l c c c c c c}
\toprule
\textbf{OOD split} & \textbf{Cond.} & $\alpha=-5$ & $\alpha=-3$ & $\alpha=0$ & $\alpha=+3$ & $\alpha=+5$ & $\Delta$ \\
\midrule
\multirow{3}{*}{JailbreakBench}
   & canonical & 0.220 & 0.200 & 0.330 & 0.480 & \textbf{0.650} & $\mathbf{+0.430}$ \\
   & native    & 0.100 & 0.190 & 0.330 & 0.420 & 0.540          & $+0.440$ \\
   & random    & 0.400 & 0.370 & 0.330 & 0.440 & 0.470          & $+0.070$ \\
\midrule
\multirow{3}{*}{XSTest}
   & canonical & 0.540 & 0.530 & 0.860 & 0.980 & \textbf{1.000} & $+0.460$ \\
   & native    & 0.390 & 0.540 & 0.860 & 0.900 & 0.940          & $+0.550$ \\
   & random    & 0.860 & 0.820 & 0.860 & 0.910 & 0.880          & $+0.020$ \\
\bottomrule
\end{tabular}

\end{table}

\section{Layer-fraction sensitivity}
\label{app:layer}

This section quantifies how detection performance depends on the choice of the per-model residual-stream layer $L_m$. We describe the layer-fraction sweep used to verify that the main result is not a brittle hyperparameter choice (\S\ref{app:layer_sweep}).

\subsection{Layer-fraction sweep}
\label{app:layer_sweep}

\paragraph{Setup.}
For each model $m$ we evaluate detection at $7$ evenly-spaced layer fractions $f \in \{1/8, 2/8, \ldots, 7/8\}$ over the model's transformer depth. The corresponding layer index is $L_m^{(f)} = \lfloor f \cdot n_{\text{layers}}^{(m)} \rceil$ (rounded). For each sweep value, we re-build the per-model anchor projector $\widehat{A}_m$ at $L_m^{(f)}$, recompute all 10 axis directions and the canonical direction, then re-run the 5-rotation detection protocol from scratch. This yields a fair apples-to-apples comparison across fractions.

The paper-main \texttt{BEST\_LAYER} dictionary (Table~\ref{tab:models}) is one specific choice; we contrast it against uniform-fraction baselines.


\begin{table}[!htbp]
\centering
\caption{Layer-fraction sensitivity (5-rotation, 5-fam mean $\pm$ std). Mid-to-late fractions (4/8 -- 7/8) all reach paper-grade performance; The per-model \texttt{BEST\_LAYER} matches uniform \(f=5/8\) within \(0.003\) binary AUROC; multi-class accuracy is also strong across the mid-layer plateau.}
\label{tab:layer_8frac}
\small
\setlength{\tabcolsep}{8pt} 
\begin{tabular}{c l c c} 
\toprule
\textbf{Fraction} $f$ & \textbf{layers (L/Q/M/P/G)} & \textbf{Multi-CLS acc} & \textbf{Binary-CLS mean} \\
\midrule
1/8 & 4 / 4 / 4 / 5 / 5     & $0.550 \pm 0.222$ & $0.830 \pm 0.087$ \\
2/8 & 8 / 7 / 8 / 10 / 10   & $0.653 \pm 0.251$ & $0.876 \pm 0.082$ \\
3/8 & 12 / 10 / 12 / 15 / 16 & $0.650 \pm 0.269$ & $0.876 \pm 0.077$ \\
\textbf{4/8} & 16 / 14 / 16 / 20 / 21 & $0.678 \pm 0.264$ & $0.900 \pm 0.085$ \\
\textbf{5/8} & 20 / 18 / 20 / 25 / 26 & $\mathbf{0.744 \pm 0.181}$ & $\mathbf{0.907 \pm 0.076}$ \\
6/8 & 24 / 21 / 24 / 30 / 32 & $0.724 \pm 0.191$ & $0.902 \pm 0.084$ \\
7/8 & 28 / 24 / 28 / 35 / 37 & $0.695 \pm 0.222$ & $0.893 \pm 0.087$ \\
\bottomrule
\end{tabular}
\vspace{-1em}
\end{table}

\section{Anchor-pool and source-pool sensitivity}
\label{app:anchor_source_sensitivity}

This section provides full per-configuration tables for the two ablations summarized in the main paper sensitivity figure (Section~\ref{sec:sensitivity}): (i) varying the anchor count $k$ per HELM scenario, and (ii) varying the source count $N$ in the LQPG $\rightarrow$ Mistral transfer pool. Both ablations use Mistral-7B as the unseen target and the LogRegression-trained probe as the classifier.

\subsection{Anchor count $k$ per HELM scenario}
\label{app:anchor_count}

We sweep the per-scenario anchor count $k \in \{5, 10, 20, 30, 40\}$, with the total pool size $N_{\mathrm{anchor}} = 15 k$ since 15 HELM scenarios contribute to the pool. For each $k$, we re-sample the pool from scratch (seed 42) and rebuild all per-model anchor projectors $\widehat{A}_m$. The same axes and the same LQPG $\rightarrow$ Mistral split are used throughout.

\begin{table}[!htbp]
\centering
\small
\setlength{\tabcolsep}{6pt} 
\begin{tabular}{c c c c c} 
\toprule
\textbf{$k$} & \textbf{$N_{\mathrm{anchor}}$} & \textbf{Multi-CLS (acc)} & \textbf{Binary-CLS (mean AUROC)} & \textbf{notes} \\
\midrule
  5  & 75  & 0.832 & 0.952 & \\
  10 & 150 & 0.834 & 0.954 & \\
  20 & 300 & 0.838 & 0.956 & paper main \\
  30 & 450 & 0.837 & 0.958 & \\
  40 & 600 & 0.843 & 0.959 & \\
\bottomrule
\end{tabular}
\caption{Anchor-count $k$ sensitivity (Mistral-unseen, LQPG sources). Even the smallest pool $k{=}5$ ($N_{\mathrm{anchor}}{=}75$) reaches $0.832$ Multi-CLS / $0.952$ Binary-CLS.}
\label{tab:anchor_count_full}
\end{table}

\paragraph{Interpretation.}
The anchor pool is robust well below paper-main size. This makes ACS construction practical even for very large unseen models where forwarding 600 anchors is expensive.

\newpage
\subsection{Source count $N$, full per-combination breakdown}
\label{app:source_count}

For source count $N$, the LQPG pool yields $\binom{4}{N}$ combinations: $4$ singletons ($N{=}1$), $6$ pairs, $4$ triples, and the single full pool ($N{=}4$). Each row below is one combination. The mean $\pm$ std rows summarize each set; these are what is plotted in the main sensitivity figure.

\begin{table}[!htbp]
\centering
\small
\setlength{\tabcolsep}{8pt}
\begin{tabular}{c l c c} 
\toprule
\textbf{$N$} & \textbf{Sources} & \textbf{Multi-CLS (acc)} & \textbf{Binary-CLS (mean AUROC)} \\
\midrule
1 & L                   & 0.799 & 0.938 \\
1 & Q                   & 0.799 & 0.951 \\
1 & P                   & 0.775 & 0.943 \\
1 & G                   & 0.472 & 0.922 \\
  & \textit{mean over 4 combos} & $0.711 \pm 0.138$ & $0.938 \pm 0.011$ \\
\midrule
2 & L+Q                & 0.875 & 0.954 \\
2 & L+P                & 0.829 & 0.950 \\
2 & L+G                & 0.852 & 0.943 \\
2 & Q+P                & 0.813 & 0.952 \\
2 & Q+G                & 0.803 & 0.951 \\
2 & P+G                & 0.789 & 0.945 \\
  & \textit{mean over 6 combos} & $0.827 \pm 0.029$ & $0.949 \pm 0.004$ \\
\midrule
3 & L+Q+P              & 0.841 & 0.955 \\
3 & L+Q+G              & 0.856 & 0.954 \\
3 & L+P+G              & 0.826 & 0.952 \\
3 & Q+P+G              & 0.812 & 0.953 \\
  & \textit{mean over 4 combos} & $0.834 \pm 0.016$ & $0.954 \pm 0.001$ \\
\midrule
4 & L+Q+P+G            & 0.838 & 0.956 \\
  & \textit{mean over 1 combo}  & $0.838 \pm 0.000$ & $0.956 \pm 0.000$ \\
\bottomrule
\end{tabular}
\caption{Source-count $N$ ablation, full per-combination breakdown (Mistral-unseen, LQPG candidate pool). The mean rows aggregate over each combination set and are what is plotted in the sensitivity figure.}
\label{tab:source_count_full}
\end{table}

\paragraph{Interpretation.}
These per-combination numbers show that ACS-based transfer is robust to source-pool composition.

\begin{itemize}\item \textbf{Single-source baselines are already strong} for the LQP family members: Llama-only and Qwen-only reach $0.799$ Multi-CLS, only $0.04$ below the full LQPG pool ($0.838$); Phi-only is $0.775$. Gemma-only collapses to $0.472$, dragging the $N{=}1$ mean down to $0.711 \pm 0.138$.\item \textbf{Adding a second source gives a $\sim+0.12$ jump} in mean Multi-CLS ($0.711 \to 0.827$) but standard deviation also drops nearly $5\times$ ($0.138 \to 0.029$), driven mostly by averaging out the Gemma-only outlier.

\item \textbf{$N{=}3$ and $N{=}4$ are essentially identical} (Multi-CLS $0.834$ vs $0.838$, Binary-CLS $0.954$ vs $0.956$). Any drop-one-out triple from LQPG matches the full LQPG pool within standard deviation.

\item \textbf{The Gemma-included triples track the rest within $\pm 0.03$ Multi-CLS}: L+Q+G ($0.856$), L+P+G ($0.826$), Q+P+G ($0.812$) versus L+Q+P ($0.841$). Once the canonical direction is computed from $\geq 3$ sources, Gemma's misaligned axes (factual, science) get washed out by averaging.
\end{itemize}

These per-combination numbers also clarify the failure mode of Gemma-only sourcing ($N{=}1$, Multi-CLS $0.472$): Multi-CLS is a 10-way classifier, and with a single Gemma source, the directions for factual / science axes are inconsistent with the Mistral-target geometry, causing systematic mis-classification on those axes.

\newpage

\section{Universal similarity supplementary}
\label{app:universal_similarity}

This section gives the per-axis breakdown behind the heatmap and PCA scatter in Section~\ref{sec:universal_similarity}, plus a 10-axis t-SNE embedding that supplements the 7-axis main-text figure.

\subsection{Per-axis cossim breakdown across model pairs}
\label{app:per_axis_cossim}

The 5$\times$5 heatmap reports the axis-mean cosine similarity between per-model anchor-projected axis directions, one cell per model pair. Table~\ref{tab:cossim_per_axis_full} disaggregates this average into one row per behavioral axis and one column per model pair, exposing where the cluster behavior comes from and where Gemma's anti-alignment originates.We use the upper-triangular ordering of the LQMPG model pool, giving 10 distinct pairs: 6 LQMP-only pairs and 4 Gemma pairs.

\begin{table}[!htbp]
\centering
\small
\setlength{\tabcolsep}{3pt}
\begin{tabular}{l c c c c c c c c c c c}
\toprule
\textbf{Axis} & \textbf{L-Q} & \textbf{L-M} & \textbf{L-P} & \textbf{L-G} & \textbf{Q-M} & \textbf{Q-P} & \textbf{Q-G} & \textbf{M-P} & \textbf{M-G} & \textbf{P-G} & \textbf{mean} \\
\midrule
\texttt{refusal}    & $\mathbf{+0.90}$ & $\mathbf{+0.90}$ & $\mathbf{+0.87}$ & $+0.26$ & $\mathbf{+0.93}$ & $\mathbf{+0.92}$ & $+0.23$ & $\mathbf{+0.94}$ & $+0.30$ & $+0.29$ & $+0.65$ \\
\texttt{math}       & $\mathbf{+0.93}$ & $\mathbf{+0.91}$ & $\mathbf{+0.91}$ & $\mathbf{+0.62}$ & $\mathbf{+0.95}$ & $\mathbf{+0.97}$ & $\mathbf{+0.62}$ & $\mathbf{+0.95}$ & $\mathbf{+0.63}$ & $\mathbf{+0.63}$ & $+0.81$ \\
\texttt{science}    & $+0.45$ & $\mathbf{+0.52}$ & $+0.41$ & $-0.11$ & $\mathbf{+0.72}$ & $\mathbf{+0.65}$ & $-0.09$ & $\mathbf{+0.66}$ & $-0.07$ & $-0.03$ & $+0.31$ \\
\texttt{factual}    & $\mathbf{+0.80}$ & $\mathbf{+0.71}$ & $\mathbf{+0.71}$ & $-0.11$ & $\mathbf{+0.90}$ & $\mathbf{+0.90}$ & $-0.14$ & $\mathbf{+0.93}$ & $-0.11$ & $-0.13$ & $+0.45$ \\
\texttt{sycophancy} & $\mathbf{+0.82}$ & $\mathbf{+0.89}$ & $\mathbf{+0.75}$ & $+0.20$ & $\mathbf{+0.88}$ & $\mathbf{+0.86}$ & $+0.08$ & $\mathbf{+0.87}$ & $+0.15$ & $+0.04$ & $+0.55$ \\
\texttt{toxicity}   & $\mathbf{+0.86}$ & $\mathbf{+0.91}$ & $\mathbf{+0.90}$ & $+0.23$ & $\mathbf{+0.87}$ & $\mathbf{+0.91}$ & $+0.24$ & $\mathbf{+0.94}$ & $+0.21$ & $+0.22$ & $+0.63$ \\
\texttt{sentiment}  & $\mathbf{+0.65}$ & $\mathbf{+0.59}$ & $\mathbf{+0.71}$ & $+0.03$ & $\mathbf{+0.82}$ & $\mathbf{+0.87}$ & $+0.15$ & $\mathbf{+0.90}$ & $+0.12$ & $+0.13$ & $+0.49$ \\
\texttt{emotion}    & $\mathbf{+0.86}$ & $\mathbf{+0.79}$ & $\mathbf{+0.87}$ & $+0.27$ & $\mathbf{+0.87}$ & $\mathbf{+0.91}$ & $+0.23$ & $\mathbf{+0.80}$ & $+0.23$ & $+0.27$ & $+0.61$ \\
\texttt{bias-G}     & $\mathbf{+0.74}$ & $\mathbf{+0.56}$ & $+0.34$ & $+0.22$ & $\mathbf{+0.79}$ & $\mathbf{+0.71}$ & $+0.36$ & $\mathbf{+0.75}$ & $+0.32$ & $+0.32$ & $+0.51$ \\
\texttt{bias-R}     & $\mathbf{+0.77}$ & $\mathbf{+0.57}$ & $+0.42$ & $+0.11$ & $\mathbf{+0.79}$ & $\mathbf{+0.73}$ & $+0.22$ & $\mathbf{+0.79}$ & $+0.22$ & $+0.24$ & $+0.49$ \\
\midrule
\textsc{axis-mean}  & $+0.78$ & $+0.74$ & $+0.69$ & $+0.17$ & $+0.85$ & $+0.84$ & $+0.19$ & $+0.85$ & $+0.20$ & $+0.20$ & $+0.55$ \\
\bottomrule
\end{tabular}
\caption{Per-axis cosine similarity between canonical directions in ACS, for all 10 LQMPG model pairs. Boldface marks values $\geq +0.5$. The bottom \textsc{axis-mean} row aggregates over all 10 axes and corresponds to the cell values of the 5$\times$5 heatmap reported in Section~\ref{sec:universal_similarity}; the rightmost column aggregates over all 10 pairs and gives a per-axis robustness measure.}
\label{tab:cossim_per_axis_full}
\end{table}

\paragraph{Analysis.}
\begin{itemize}
    \item  Most LQMP-only pairs are strongly aligned across most axes, with the strongest consistency in the Q-M, Q-P, and M-P pairs. 
     The main exceptions are science and the two bias axes for some Llama/Phi-related pairs.
    \item \textbf{Gemma-pair (L-G, Q-G, M-G, P-G) cosines split into two regimes.} On \texttt{math} they remain solidly positive ($+0.62$--$+0.63$), behaving like the LQMP cluster. On \texttt{factual} and \texttt{science} they go negative ($-0.07$ to $-0.14$), indicating that Gemma's contrastive direction for these axes points the opposite way relative to the LQMP cluster. On the remaining axes (refusal, toxicity, sentiment, sycophancy, emotion, bias) Gemma pairs sit in the $[+0.04, +0.32]$ range -- positive but weak.
    \item \textbf{The two bias axes (bias-G, bias-R) behave nearly identically across pairs}. This is consistent with the 10-way bias-cluster collapse documented in detection.
\end{itemize}

\subsection{10-axis t-SNE embedding}
\label{app:tsne_e}

\begin{figure}[!htbp]
\centering
\includegraphics[width=0.8\linewidth]{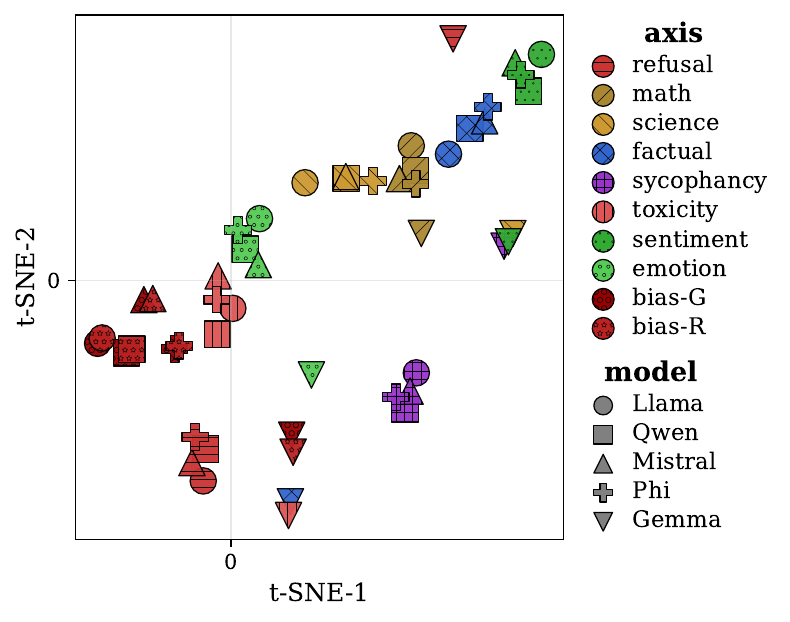}
\caption{t-SNE embedding of the 50 per-model anchor-projected axis directions (\(5\) models \(\times\) \(10\) axes) in ACS. Hatch pattern encodes axis identity, marker shape encodes model family. Same-axis points cluster tightly across LQMP (4-fam) on most axes; Gemma sits at the periphery for \texttt{toxicity}/\texttt{emotion} and flips to a separate sub-cluster for \texttt{factual} and \texttt{science}. The two bias axes (\texttt{bias-G}, \texttt{bias-R}) overlap heavily, supporting the single-bias-cluster reading in detection.}
\label{fig:tsne_10axis}
\end{figure}

The main-text PCA / t-SNE figure (Section~\ref{sec:universal_similarity}) uses a 7-axis subset that excludes the two bias axes and emotion. For completeness, Figure~\ref{fig:tsne_10axis} shows the same t-SNE embedding extended to all 10 axes. The cluster structure is preserved: each axis forms a tight cluster across LQMP, with Gemma either sitting at the periphery or detaching entirely on \texttt{factual} and \texttt{science}. The two bias axes form a single mixed cluster, consistent with the per-axis cossim table above and with the bias-G $\leftrightarrow$ bias-R confusion described in detection.

\end{document}